\documentclass{ecai2014}
\usepackage{times}
\usepackage{color, graphicx}
\usepackage{latexsym}
\pdfoutput=1
\usepackage[numbers]{natbib}
\usepackage[linesnumbered, ruled]{algorithm2e}
\usepackage{amssymb,amsfonts,textcomp}
\usepackage{amsthm}

\DeclareMathOperator*{\ldag}{ldag}
\DeclareMathOperator{\argmax}{argmax}

\DeclareMathOperator*{\BI}{BI}

\begin{document}

\title{How good is the Shapley value-based approach\\ to the influence maximization problem?}

\author{Kamil Adamczewski~\institute{\tiny University of Oxford, Seoul National University, email: kamil.m.adamczewski@gmail.com}  $^{~~,3}$\and Szymon Matejczyk~\institute{\tiny Institute of Computer Science, Polish Academy of Sciences, email:s.matejczyk@phd.ipipan.waw.pl}$^{~~,}$\footnote{\tiny Both first authors contributed equally to this work.}\and Tomasz P. Michalak~\institute{\tiny University of Oxford, University of Warsaw, email: tomasz.michalak@cs.ox.ac.uk}}

\vspace{-0.5cm}
\maketitle

\section{INTRODUCTION}
This paper studies the process of information diffusion over a network, where new nodes as time passes become ``informed'' (or equivalently, influenced or infected), and a related problem of influence maximization that concerns finding  the set of initial nodes that will lead to the most widespread effect of diffusion in the network ~\cite{kempe,
domrich2}.  
Influence maximization is an NP-hard problem \cite{kempe}, and therefore, literature has focused on approximating the optimal solution with greedy algorithms and various heuristics \cite{kempe, chen2010ldag, goyal2011celf++}.

In this spirit, recent research \cite{ramasuri, aadithya} proposed the Shapley
value, a concept borrowed from game theory, as a measure of node centrality. The key idea is
to define a cooperative game over a network in which players are nodes,
coalitions are groups of nodes, and payoffs of coalitions depend on how many nodes these coalitions can infect as a group. 
Next, a game-theoretic solution concept---most often the Shapley value---quantifies a role or importance of a player (e.g. in the diffusion process) by its average marginal contribution in the game. In this paper, we also briefly test the Banzhaf index (BI) as an alternative solution concept.

Nevertheless, unlike other centrality measures, the game-theoretic centrality for influence maximization has not been yet rigorously evaluated. The only experimental work is that of \citet{ramasuri} who proposed the SPIN algorithm built upon the Shapley value. This evaluation is, however, very preliminary, as it relies on the approximated Shapley value and is not scalable even for the graphs larger than $15K$ nodes \citet{chen2010ldag}.

In this paper, we firstly correct the above shortcomings building upon the work of \citet{aadithya}. We further show that the use of the Shapley value is not restricted to those games and  propose the use of the Shapley value in the LDAG model from \citet{chen2010ldag}. Finally, we verify the usefulness of the algorithms in \cite{aadithya} in information diffusion application and compare it against current state-of-the art algorithms for estimation of top $k$ nodes.

\section{DISCOUNT SHAPLEY VALUE CENTRALITY}

\noindent 
The importance of a node as described in the previous section can be modeled by a game where the marginal contribution of a node is the average probability that a node contributes its neighbors. A basic characteristic function can be defined as

%equation with smple fringe game 1
\begin{equation} \label{game1achar} v_1(C) = 
\begin{cases} 
0 & \mbox{if $C=\emptyset$} \\ 
\text{size}(surrounding(C)) & \mbox{otherwise} 
\end{cases} 
\end{equation}

\noindent where \textit{surrounding(C)} is the set of nodes which are the neighbours of the nodes in the coalition, formally 
$v \in surrounding(C)$ if  $\{v \in V : \exists u \in C$ such that $(u, v) \in E$ and $v \notin C\}.$

It can be easily shown that the Shapley value of this game can be computed in polynomial time by reducing this game to the Game 1 in \citet{aadithya} and it is equal to $\sum_{v_k \in \{ v \cup N(v) \}} \frac{1}{1+deg(v_k)}-1$.

In the influence maximization problem we desire to find seeds that maximize the probability that a node in a network infects its neighbors given a changing and unknown set of already infected nodes. The surrounding model addresses this issue partially and calculates the expected node contribution. However, it neglects two key aspects of information diffusion: 1) a node cannot infect itself, 2) diffusion is a process over time, not limited to one-step infection at a single point in time (thus, expected value should also be conditioned on nodes that are infected but are not part of the initial seed). These two issues are addressed in the Discounted Shapley value model. 

In resolving the first issue we take advantage of the fact that in the the above equation the Shapley value is the sum of probabilities that the node contributes each of its neighbors and itself. As we only want to consider the influence of a node on others, Algorithm ignores the probability that a given node contributes itself (lines 2-5). %The algorithm then sorts the nodes according to the diminished Shapley value (line 6).

We subsequently address the second issue, that is we attempt to account for active nodes that have been infected in the time steps $t>1$. %Given the ranking of 
%according to the diminished
%the Shapley value, 
From among the uninfected nodes we pick the node with the highest Shapley value, add it to the set of the top nodes $A$ and ``remove'' it and its neighbors from the network. %The reason behind removing the nodes is that %although the nodes with the highest Shapley value are the agents who individually can be good target nodes, when lumped together as a group of top node, they may not be a good choice for influence maximization since the top nodes may be clustered together. 
We do it because adjacent nodes are likely to share a substantial number of neighbours, meaning that when both nodes are chosen as top nodes, they will trigger influence wave in the same region of the graph, leaving more distant nodes unaffected. %We verified empirically that the desired property for top $k$ nodes is for them to be spread all over the network and does not allow neighbors to be in the initial seed. %The algorithm that we developed does not allow neighbours to be the top nodes. 
 %We then mark the neighbors of $v$ as infected and essentially ``remove'' them from the network. We then recompute the Shapley value of the nodes for the remaining part of the graph. 
 %--other option
%After removing the nodes, we recompute the Shapley value of the nodes for the remaining part of the graph.
%--
% That is, for every top node $v$ added to the set of 
Subsequently for the uninfected nodes, we update their SV by subtracting the probability that they influence ``removed'' nodes.

%\vspace{2cm}
%\SetAlTitleFnt{\tiny\sf}
%\AlCapFnt{\tiny\sf}
%\SetAlCapFnt{\tiny\sf}
\SetAlFnt{\tiny\sf}
\setlength{\textfloatsep}{4pt}% Remove \textfloatsep
%argument was "t!"
  \begin{algorithm} \label{shalgo}
  %\SetAlFnt{\tiny\sf}
 \SetAlgoLined
 \For{$i$ \KwTo $n$}{
 	\ForEach{$u \in $neighbor$(v_i)$} {
 	 shapley$[i] += \frac{1}{1+\deg(u)}$\;
  	}
 }
% sort shapley[] in a decreasing order \;
% originalshapley[] $\leftarrow$ shapley[] \;
  
  %decreaseShapley(){
 	$A \leftarrow \emptyset$; infected $\gets \emptyset$\;
  	\For{$1$ \KwTo $k$}{
  		\If{not all nodes are infected}{
  	    topnode $\gets \argmax_{i \notin \text{infected}} \{$shapley$[i]\}$ \;
  	    
			$A \leftarrow A \cup \{$topnode$\}$ \;
			infected $\gets$ infected $\cup \{$topnode$\}$\;
 		 	\ForEach{$u \in $neighbor$($topnode$)$}{
 		 	    infected $\gets$ infected $ \cup \{u\}$\;
		 		\ForEach{$i  \in $neighbor$(u)$}{
					shapley$[i]-=\frac{1}{1+\deg(u)}$;%\SetAlFnt{\tiny\sf}
				}
  			}
  		} \Else {
	Choose node $ \notin A$ with highest initial Shapley value and add to $A$\;
	}
 	}
%}	

return A containing top $k$ nodes \;
 
 \caption{\small Discounted Shapley Value}
 \end{algorithm}

\section{SHAPLEY VALUE AND BANZHAF INDEX CENTRALITY IN LOCAL DAGs}
We also propose an algorithm which incorporates  game theoretical solution concepts of the Shapley value and the Banzhaf index into the greedy approach by \cite{chen2010ldag} to find the most influential nodes called local DAG (LDAG).

The motivation behind the LDAG method comes from two observations. 1) Computing
seed spread function is $\#$P-hard under both models main models. 2) Finding set
that maximizes spread function is NP-hard even if we can compute the
spread function in polynomial time. Thus, \citet{chen2010ldag} 
propose to reduce the network and form a LDAG, a directed acyclic graph which encapsulates the approximate influence exerted on a given node in the network. LDAG is chosen for each node in order to capture as much ``influence'' from the entire network as possible. Although, \citet{chen2010ldag} prove that finding such a graph is NP-hard, they observe that a greedy algorithm performs very well in practice.
%in finding a directed acyclic graph that captures most of the influence exerted on a given
%node in a network. 

Since we assume we can reduce the entire network to a set of LDAGs, we find the initial seed set by analyzing the most influential nodes in all the LDAGs. While, in order to achieve this, \citet{chen2010ldag} use a greedy approach, we propose an alternative approach that uses the Shapley value and the Banzhaf index as measures of node centrality in the LDAGs. Since the computation of both solution concepts is usually challenging, we use Monte Carlo simulations where we approximate the solution by sampling permutations. Furthermore, we take advantage of the LDAG structure which is comparatively small compared to the size of the network. We also reduce the number of input nodes for the computation of the solution concepts (this reduced the number of necessary Monte Carlo simulations) using properties of power indices.

We particularly take advantage of the additive nature of these two solution concepts. As a result, this game-theoretic approach, as opposed to the greedy approach in \citet{chen2010ldag} is particularly suitable for distributed systems, because the resulting power indices can be computed independently on LDAGs and easily merged.  

We compute the Shapley value and the Banzhaf index assuming that the characteristic function is
the approximated influence spread in LDAG. %under both IC and LT models.
In the Banzhaf index case a node $v$ is influenced independently by its predecessors and ancestors in a given LDAG. This makes
possible to run Monte Carlo simulations for these sets independently and when calculating $\BI(v)$ in $\ldag(u)$ (DAG directed at $u$) we can forget about nodes that are neither $v$'s ancestors nor predecessors. Using this fact we can reduce the number of MC iterations even further.

\vspace{-0.4cm}
\section{EXPERIMENTS}
\label{section:experiments}
The experiments consist of two parts, 1) finding $k$ most influential nodes according to each algorithm ($k$ is 2-30\% of the network size), 2) testing the performance of the seed set by means of Monte Carlo simulations. We conduct the experiments on two diffusion models: Independent Cascade and Linear Threshold \citet{kempe} .

As far as the quality of seed set is concerned, the greedy LDAG performs consistently best across all the data sets, seed sizes and on both models (\citet{chen2010ldag} only test it on the LT model). 

The performance of CELF++ and Shapley value LDAG approach are similar on IC model, where CELF++ performs slightly better for smaller seed size and the roles reverse for larger seed size. SV LDAG performs better on the LT model which makes sense since LDAG is designed for LT model. DSV and \cite{aadithya} perform similarly in the IC model and DSV is slightly better in the LT model.
%We also note the weak performance of the \citet{aadithya} threshold model and very unstable performance of the Banzhaf index LDAG method across different datasets.
In the larger networks with thousands of nodes, the performance of the Shapley value LDAG and DSV is only preceded by the greedy LDAG. 
The three algorithms perform substantially better than the Degree Discount algorithm.

\vspace{-0.5cm}
\begin{figure}[h]
\includegraphics[trim = 0mm  2mm 0mm 2mm, clip, width=4.5cm]{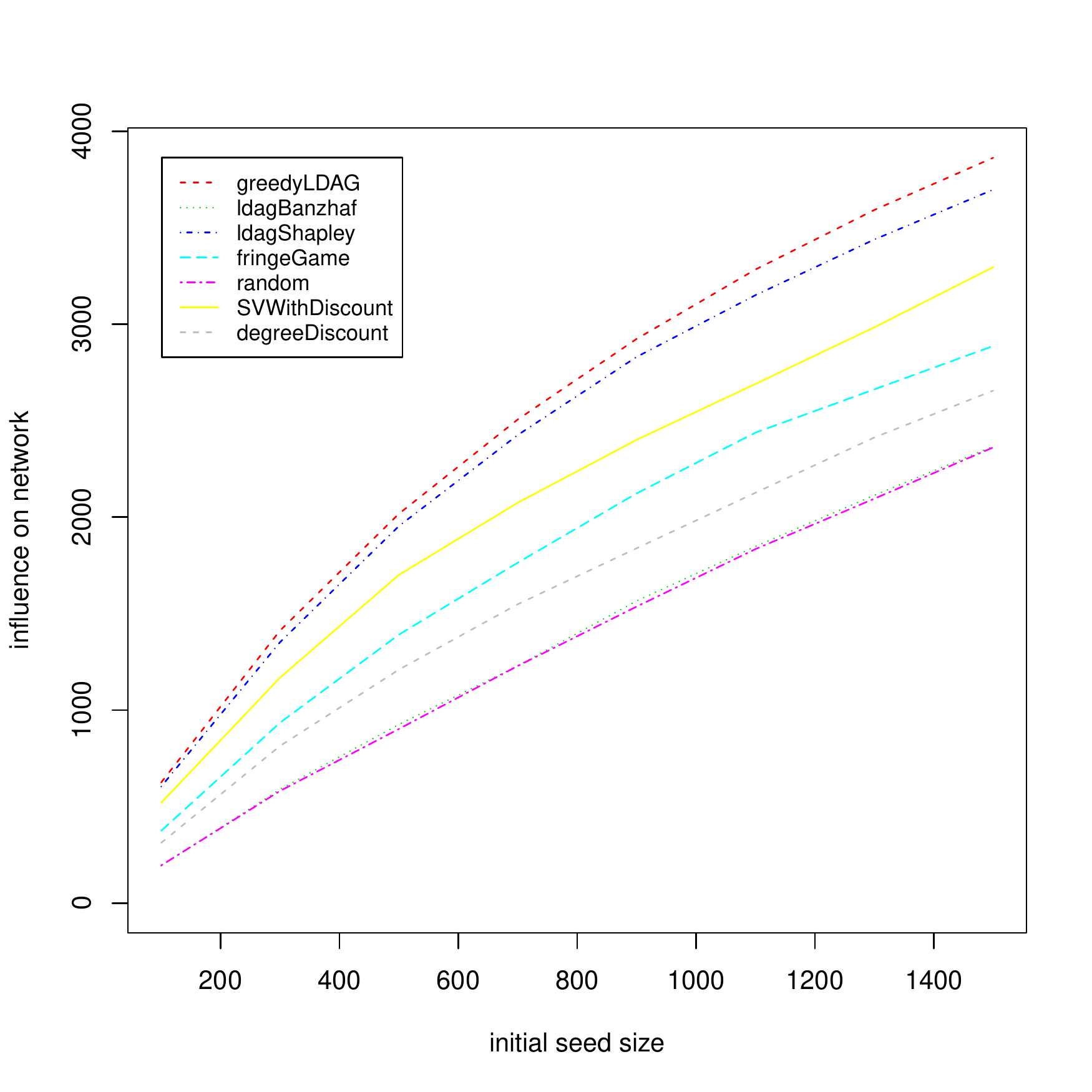}
\hfill    
\includegraphics[trim = 0mm  2mm 0mm 2mm, clip, width=4.5cm]{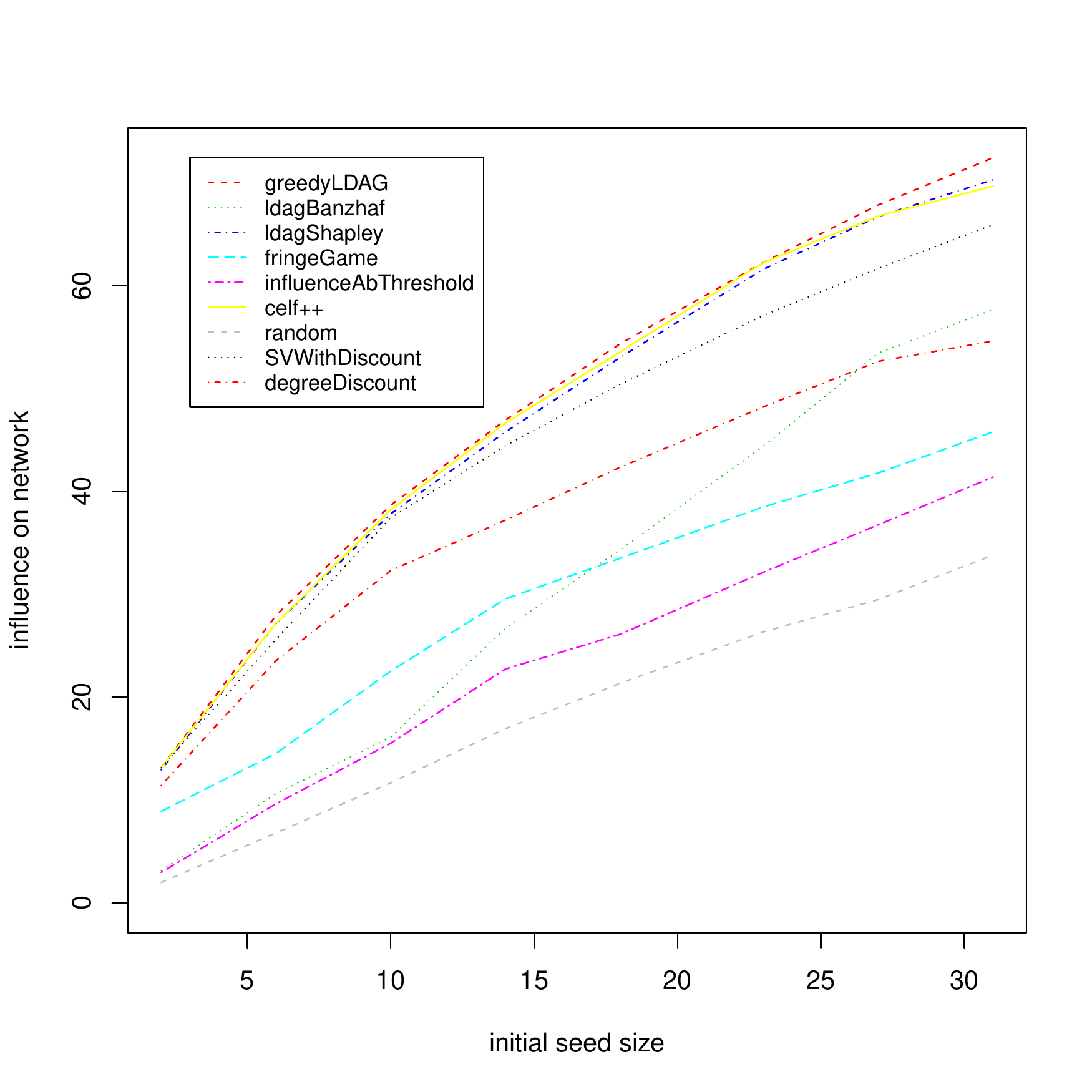}
\vspace{-0.7cm}
\caption{Comparison of various methods seed set quality (expected spread) as a function of it's size ($k$) for two different real life data.}
\end{figure}

\vspace{-0.5cm}
\section{CONCLUSION}
Our result show that the Shapley value is a competitive centrality measure for information diffusion. Specifically, we presented two algorithms that use the Shapley value in the two approaches recently proposed in the literature to determine the most influential nodes in a network: a greedy approach which relies on repeated computation of the information spread, and the heuristics which uses the Shapley value (which is exact and computable in polynomial time) as a centrality measure. We also verified the performance of the Shapley value-based centrality proposed in the work of \citet{aadithya}. The experimental result show that the greedy LDAG approach comes up with the highest quality seed set. Yet our proposed heuristic based on the Shapley value performs almost as good as the greedy algorithm in terms of solution quality, and it can be easy adopted to Map-Reduce scheme. Finally, the Discount Shapley value centrality heuristic performs better than the models from \citet{aadithya} and the current state-of-the-art  Discount Degree heuristic; thus, narrowing the gap between the centrality based heuristics and greedy approximations.

\vspace{-0.4cm}
\footnotesize{
\ack{Kamil Adamczewski \& Tomasz Michalak were supported by the Polish National Science Centre grant DEC-2013/09/D/ST6/03920 (University of Warsaw). Szymon Matejczyk was supported by the European Union from resources of the European Social Fund. Project PO KL ``Information technologies: Research and their interdisciplinary applications''. Tomasz Michalak was supported by the European Research Council under Advanced Grant 291528 (“RACE”) at the University of Oxford.}
}

\vspace{-0.2cm}
{\footnotesize
\bibliographystyle{plainnat}
\bibliography{ECAI-57}
}

\end{document}